\documentclass{article} %
\usepackage{iclr2026_conference,times}

\usepackage{amsmath,amsfonts,bm}

\def\eqref#1{equation~\ref{#1}}

\def\1{\bm{1}}

\DeclareMathAlphabet{\mathsfit}{\encodingdefault}{\sfdefault}{m}{sl}
\SetMathAlphabet{\mathsfit}{bold}{\encodingdefault}{\sfdefault}{bx}{n}

\usepackage{amsmath,amsfonts}
\usepackage[ruled,linesnumbered]{algorithm2e}
\usepackage{array}
\usepackage[caption=false,font=normalsize,labelfont=sf,textfont=sf]{subfig}
\usepackage{textcomp}
\usepackage{stfloats}
\usepackage{url}
\usepackage{verbatim}

\usepackage{graphicx}
\usepackage{amssymb}
\usepackage{booktabs}
\usepackage{times}
\usepackage{microtype}
\usepackage{epsfig}
\usepackage{caption}
\usepackage{float}
\usepackage{placeins}
\usepackage{color, colortbl}
\usepackage{enumitem}
\usepackage{tabularx}
\usepackage{xstring}
\usepackage{multirow}
\usepackage{xspace}
\usepackage{xcolor}
\usepackage[hang,flushmargin,symbol]{footmisc}

\usepackage{threeparttable}

\usepackage{lipsum}

\usepackage{marvosym}

\usepackage{hyperref}

\usepackage{makecell}

\usepackage{wrapfig}

\usepackage[normalem]{ulem}

\definecolor{softgreen}{rgb}{0.0, 0.6, 0.2}
\definecolor{softblue}{rgb}{0.8, 0.9, 1.0}

\DefineFNsymbols{lettersymbol}{\Letter \dagger \ddagger \S \P | {**}} \setfnsymbol{lettersymbol}

\title{Interleaving Reasoning\\ for Better Text-to-Image Generation}

\iclrfinalcopy
\author{
    Wenxuan Huang$^{1,2}$,
    Shuang Chen$^4$,
    Zheyong Xie$^3$,
    Shaosheng Cao$^3$\thanks{Corresponding authors.}\quad,
    Shixiang Tang$^2$, \\
    \quad\textbf{Yufan Shen}$^5$,
    \textbf{Qingyu Yin}$^5$,
    \textbf{Wenbo Hu}$^4$,
    \textbf{Xiaoman Wang}$^1$,
    \textbf{Yuntian Tang}$^1$,
    \textbf{Junbo Qiao}$^1$, \\
    \quad\textbf{Yue Guo}$^4$,
    \textbf{Yao Hu}$^3$,
    \textbf{Zhenfei Yin}$^{6}$\footnotemark[1]\quad,
    \textbf{Philip Torr}$^6$,
    \textbf{Yu Cheng}$^2$,
    \textbf{Wanli Ouyang}$^2$,
    \textbf{Shaohui Lin}$^{1}$\footnotemark[1]\quad \\
    {\normalsize$^1$East China Normal University} \quad
    {\normalsize$^2$The Chinese University of Hong Kong} \quad 
    {\normalsize$^3$Xiaohongshu Inc.} \\
    {\normalsize$^4$University of California, Los Angeles} \quad 
    {\normalsize$^5$Zhejiang University} \quad
    {\normalsize$^6$University of Oxford} \\
    {\tt osilly0616@gmail.com} \\
    Github Repo: \url{https://github.com/Osilly/Interleaving-Reasoning-Generation}
}

\begin{document}

\maketitle

\begin{figure}[H]
    \centering
    \includegraphics[scale = 0.23]{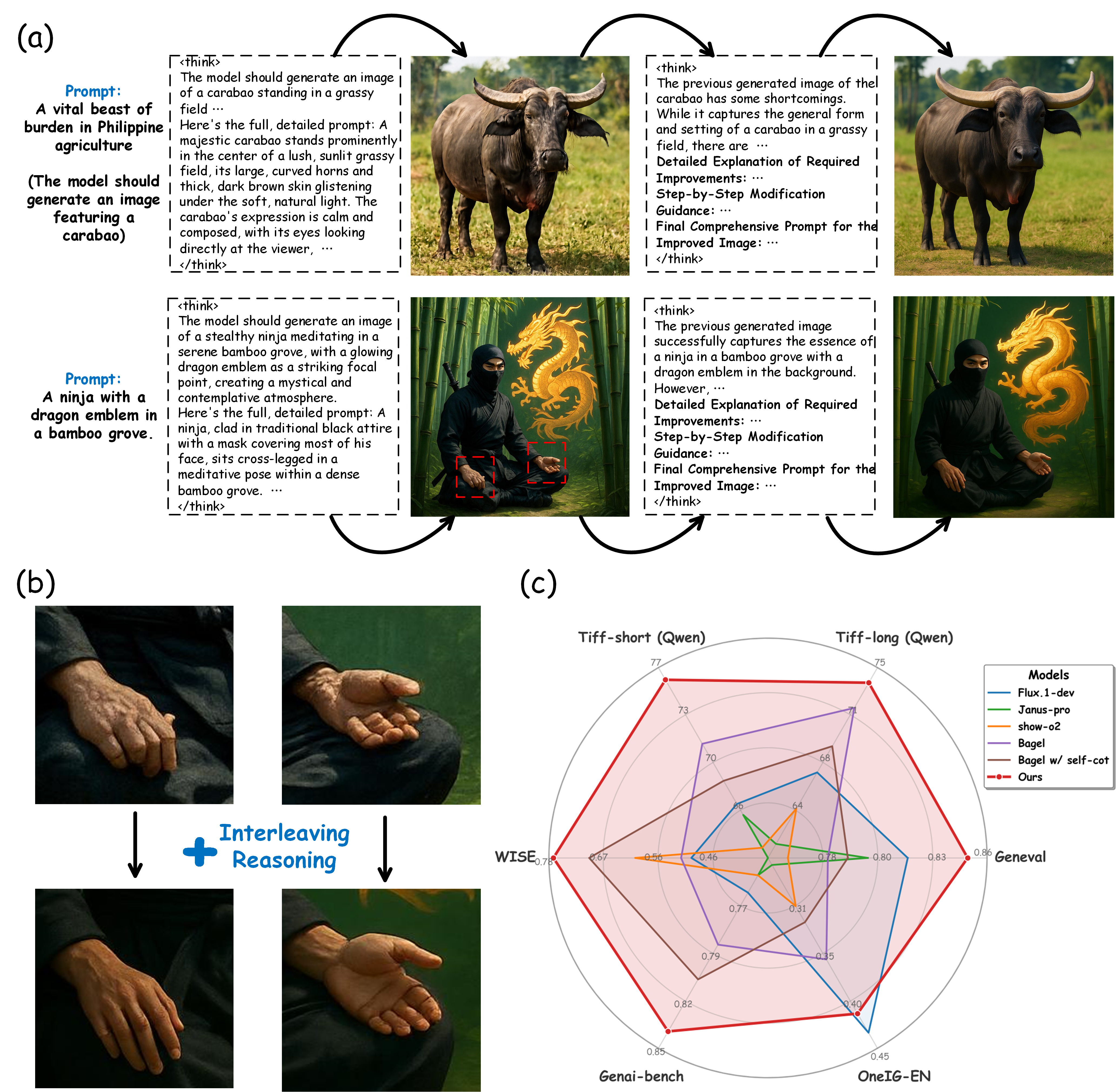}
    \caption{
        As shown in (a), we illustrate an example of Interleaving Reasoning Generation (IRG). 
        Given a prompt, the model first produces a text‑based reasoning process and then generates an image conditioned on that reasoning. Next, building upon the initial image, the model reflects on how to improve its quality and produces a refined image through this reflection process.
        IRG can substantially enhance image generation quality. 
        For instance, in the top case of (a), IRG improves upon the previous generated image via multi‑turn reasoning, enhancing rendering textures, shadow realism, and other visual properties. 
        In the bottom case of (a), IRG significantly improves fine‑grained details, such as the delicate structures of fingers—highlighted within the red box (as detailed in (b)).
        As shown in (c), compared to current SoTA models, our proposed IRG achieves clearly superior performance across multiple mainstream T2I benchmarks.
    }
    \label{fig:case}
\end{figure}

\begin{figure}[H]
    \centering
    \includegraphics[scale = 0.43]{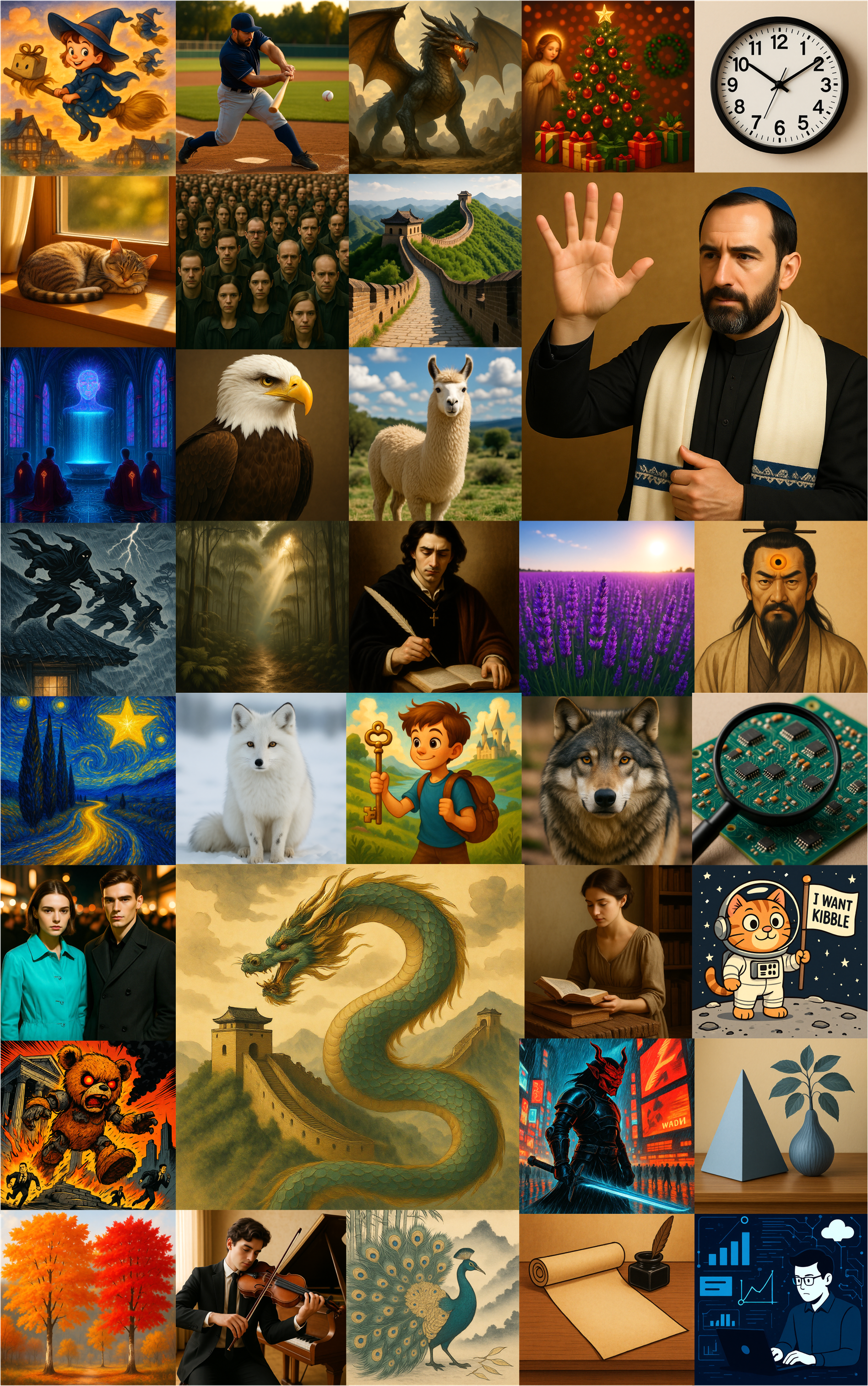}
    \caption{
        Visualization results of IRG at 1024×1024 resolution.
        The examples are selected from WISE~\citep{niu2025wise}, TIIF~\citep{wei2025tiifbenchdoest2imodel}, and GenAI-Bench~\citep{li2024genaibenchevaluatingimprovingcompositional}.
    }
    \label{fig:big_case}
\end{figure}

\begin{abstract}
Unified multimodal understanding and generation models recently have achieve significant improvement in image generation capability, yet a large gap remains in instruction following and detail preservation compared to systems that tightly couple comprehension with generation such as GPT-4o.
Motivated by recent advances in interleaving reasoning, we explore whether such reasoning can further improve Text-to-Image (T2I) generation. 
We introduce \textbf{\textit{Interleaving Reasoning Generation (IRG)}}, a framework that alternates between text-based thinking and image synthesis: the model first produces a text-based thinking to guide an initial image, then reflects on the result to refine fine-grained details, visual quality, and aesthetics while preserving semantics. 
To train IRG effectively, we propose \textbf{\textit{Interleaving Reasoning Generation Learning (IRGL)}}, which targets two sub-goals: (1) strengthening the initial think-and-generate stage to establish core content and base quality, and (2) enabling high-quality textual reflection and faithful implementation of those refinements in a subsequent image. 
We curate \textbf{\textit{IRGL-300K}}, a 300K-scale dataset organized into six decomposed learning modes that jointly cover learning text-based thinking, and full thinking–image trajectories. 
Starting from a unified foundation model that natively emits interleaved text–image outputs, our two-stage training first builds robust thinking and reflection, then efficiently tunes the IRG pipeline in the full thinking–image trajectory data. 
Extensive experiments show SoTA performance, yielding absolute gains of \textbf{5–10 points} on GenEval, WISE, TIIF, GenAI-Bench, and OneIG-EN, alongside substantial improvements in visual quality and fine-grained fidelity. 
As an early exploration, our results demonstrate that interleaving reasoning is a powerful paradigm for advancing T2I.
The code, model weights and datasets will be released in: \url{https://github.com/Osilly/Interleaving-Reasoning-Generation}.
\end{abstract}

\section{Introduction}

Unified multimodal understanding and generation models consolidate image/text understanding and image synthesis capabilities within a single foundation model, and have recently emerged as a focal point of interest in the research community~\citep{sun2024emu,chameleon,tong2024metamorph,wu2025janus,xie2024show,chen2025blip3,xiao2025mindomni,xiao2025omnigen,liao2025mogao,xie2025show,wu2025omnigen2,deng2025emerging}.
Representative efforts in this line of research, exemplified by GPT‑4o~\citep{openai2025chatgpt4o}, seamlessly integrate comprehension and generation capabilities. 
This enables a pronounced performance gap relative to existing unified models, particularly in instruction-following for image generation and in the preservation of visual details~\citep{deng2025emerging}.

Motivated by recent advances in text-based reasoning, notably test-time scaling techniques for (Multimodal) Large Language Models ((M)LLMs)~\citep{jaech2024openai,guo2025deepseek,huang2025vision, chen2025advancing}, a growing body of work has explored whether incorporating such text-based reasoning processes can yield improvements in the fidelity and overall quality of image generation~\citep{fang2025got,xiao2025mindomni,deng2025emerging,jiang2025t2i}.
These works underscore this perspective and seek to exploit large-scale interleaved text–image corpora to learn subtle cross-modal interaction patterns, thus enabling seamless knowledge transfer between the understanding and generation stages of the model.
However, in Text-to-Image (T2I) tasks, they employ only a single textual segment as auxiliary supervision in T2I generation, with the objective of producing outputs that more faithfully adhere to the original prompt.
Recently, some works in (M)LLM field have focused on the \textit{interleaving reasoning}, \textit{i.e.}, multi-turn interactions and exhibits sophisticated reasoning dynamics~\citep{interleaving-reasoning}, while this reasoning modality has empirical demonstrated superior accuracy in addressing complex problems~\citep{openai-o3-and-o4-mini,openai-deep-research}.
This observation motivates the exploration:

\textit{Whether interleaving reasoning can further enhance T2I generation quality?}

As shown in Fig~\ref{fig:case}, in generating a prompt‑aligned image, the model is typically able to produce content that is broadly correct in terms of semantics, yet it remains challenging to attain superior visual quality and fine‑grained fidelity (\textit{e.g.}, in rendering textures, shadow realism, and delicate structures such as fingers).
The idea is intuitive: if high‑quality image synthesis is considered a hard problem, adopting a multi‑step reasoning strategy to tackle it is both reasonable and necessary.
To solve this, a straightforward idea is to have the model first produce a text-based thinking process and generate an image based on that reasoning. 
Then, building on the initial image, the model reflects on how to improve its quality, and produces an improved image through reflection.
We denote this process as \textbf{\textit{Interleaving Reasoning Generation (IRG)}}.
Thus, we argue two points, 1) an additional text‑based reasoning process can serve as auxiliary supervision for image generation, thereby alleviating the difficulty of direct generation, and 2) producing one image that simultaneously attains high visual quality and precise instruction following is non‑trivial, whereas a multi‑turn generation strategy can incrementally refine the output toward the desired goal.
While these positions are aligned with prior reflection-based T2I generation approaches, the key distinction lies in their goals: prior methods~\citep{zhuo2025reflection,wu2025omnigen2,chern2025thinking} generally employ reflection to rectify major semantic or structural errors in the generated content, with some adopting non‑end‑to‑end frameworks, whereas our approach focuses on leveraging reflection to refine fine-grained details and improve overall visual quality in an end‑to‑end manner, with the main subject matter established during the initial generation.
Specifically, through IRG, we not only enhance the semantic correctness of the generated content, but also focus on improving the quality, fine‑grained details, and aesthetic aspects of the generated images.

Based on the insights, we firstly select a unified multimodal understanding and generation model as the base model, given its capability to produce interleaved text–image outputs.
To facilitate interleaving reasoning-based generation with reflection, we propose the \textbf{\textit{Interleaving Reasoning Generation Learning (IRGL)}} paradigm and formulate the objective as two sub-goals.
The first is to strengthen the model’s initial thinking and generation stage, which establishes the core content and base quality of the generated image.
The second is to equip the model with the ability to produce detailed, high-quality text-based reflections, and to generate enhanced images that faithfully implement those reflections.
In particular, we propose the \textbf{\textit{IRGL-300K}} dataset, which refines the aforementioned two objectives into two complementary focuses: learning text-based thinking and mastering the image generation with thinking pipeline. 
It comprises \textit{six decomposed learning modes} that jointly target comprehensive enhancement of model performance throughout the IRG process, while ensuring optimal exploitation of data resources.
In the training pipeline, the model is first trained to generate accurate initial thinking given a prompt, as well as to produce text-based reflections based on the initial reasoning step that improve quality. 
Furthermore, we incorporate full thinking–image trajectories to mitigate potential degradation in the model’s core generative ability in this training stage.
Then, leveraging the acquired thinking generation capability, we utilize the full thinking–image trajectory data to efficiently train the entire IRG pipeline.

Extensive experiments demonstrate that our proposed IRG achieves State‑of‑The‑Art (SoTA) benchmark performance, delivering absolute improvements of 5–10 points across multiple benchmarks, including GenEval, WISE, TIIF, GenAI‑Bench, and OneIG‑EN. 
In addition, IRG significantly enhances visual quality and fine‑grained fidelity. 
As an early effort to introduce Interleaving Reasoning into the T2I domain, we hope our work can inspire future research in this direction.

\begin{figure}[t]
    \centering
    \includegraphics[scale = 0.14]{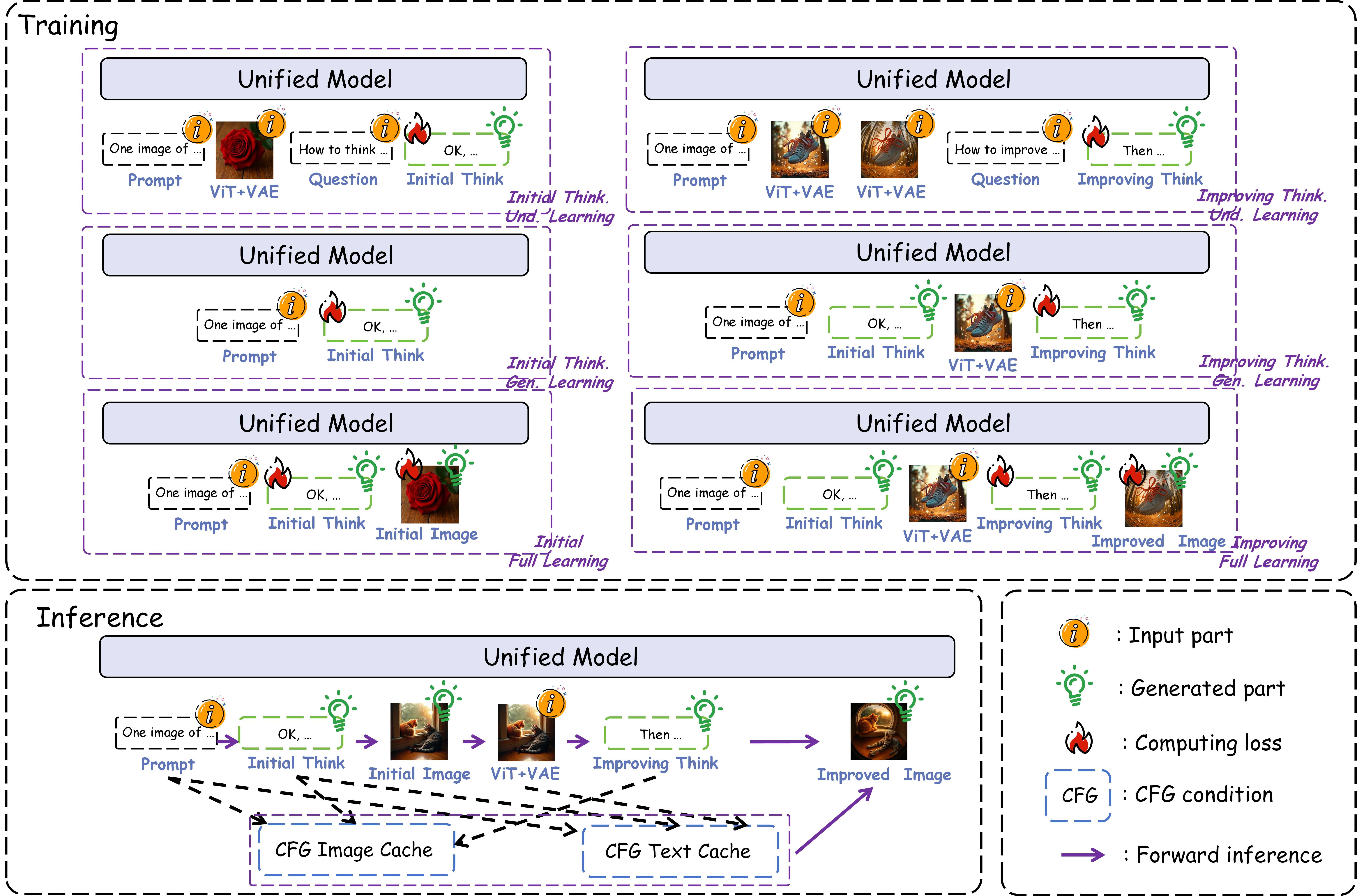}
    \caption{
        Overview of our proposed IRG training and inference pipeline.
        IRG learns the text-based thinking process and the complete high-quality image generation pipeline under six decomposed learning modes.
        During inference, we introduce a dedicated CFG condition design~\citep{ho2022classifier} for IRG’s improved image generation steps. 
    }
    \label{fig:overview}
\end{figure}

\section{Method}

\subsection{Preliminaries and Notations}

Unified multimodal understanding and generation models based on an integrated transformer architecture, \textit{e.g.}, BAGEL~\citep{deng2025emerging}, jointly perform image understanding and generation within a single architecture, facilitated by large‑scale training on interleaved text–image data. 
Such models can naturally handle both interleaved inputs and interleaved outputs.
Let $T_{in}$ and $I_{in}$ represent the input text and image, respectively, while $T_{out}^{(1)}$ and $I_{out}^{(1)}$ denote the initial output text and image, respectively.
The standard image understanding and T2I generation with self‑CoT reasoning process can be formulated as:
\begin{equation}
    \begin{split}
    & T_{in}+I_{in}\rightarrow T_{out}^{(1)}, \\
    & T_{in}\rightarrow T_{out}^{(1)}\rightarrow I_{out}^{(1)}.
    \end{split}
    \label{eq1}
\end{equation}
Previous works that adopt self‑CoT~\citep{deng2025emerging,fang2025got,xiao2025mindomni,jiang2025t2i} typically focus solely on text‑based reasoning to improve image generation, while overlooking the potential of leveraging the initially generated image to further enhance visual quality and perform multi‑step information fusion for better results.
Thus, to conduct additional T2I reasoning step conditioned on the previously generated text and image, we first encode the generated image into Vision Transformer (ViT) features and Variational Autoencoder (VAE) features, which are defined as $I_{f}$.
Then, based on the full information cache, the multi-turn generation can be conducted and the pipeline denoted as follows:
\begin{equation}
    \begin{split}
    & T_{in}
    \rightarrow T_{out}^{(1)} \rightarrow I_{out}^{(1)} \xrightarrow{\mathrm{enc}} I_{f}^{(1)}
    \rightarrow T_{out}^{(2)} \rightarrow I_{out}^{(2)} \xrightarrow{\mathrm{enc}} I_{f}^{(2)}
    \rightarrow \cdots 
    \rightarrow T_{out}^{(n)} \rightarrow I_{out}^{(n)},
    \end{split}
\label{eq2}
\end{equation}
where $n$ means the total turn number and $\xrightarrow{\mathrm{enc}}$ represents the image feature encoding process, and note that for intermediate images generated at the $k$‑th turn (\textit{e.g.}, $I_{\text{out}}^{(k)}$), only their encoded representations $I_{f}^{(k)}$ are propagated to the subsequent computation stages.
During the generation of the final image $I_{out}^{(n)}$, the model exchanges and exploits multiple segments of interleaved text–image representations, a process we term \textbf{\textit{Interleaving Reasoning Generation (IRG)}}. 
By formulating the synthesis of the final image as the ultimate goal, the model employs multi‑stage progressive generation and cross‑modal information fusion to maximize output quality.
We note that an inherent strength of unified models lies in their capacity to process and generate interleaved inputs and outputs, inherently supporting multi‑turn generation with self‑CoT reasoning. 
This property enables us to investigate how interleaving reasoning within the generation pipeline can further extend the achievable limits of generative capability.

\subsection{Interleaving Reasoning Generation}

\subsubsection{Overview}

As mentioned above, IRG can be defined as comprising two components: (1) an initial text‑based reasoning process followed by image generation based on that reasoning, and (2) repetition of the first component to produce an improved image. 
In this work, we focus solely on a single refinement iteration, \textit{i.e.}, limiting the second component to one turn (set $n$ in Eq~\ref{eq2} to 2), in order to validate our hypothesis on whether Interleaving Reasoning can effectively enhance text‑to‑image generation quality.

In the following sections, we detail the \textit{\textbf{Interleaving Reasoning Generation Learning (IRGL)}} framework and explain how different forms of interleaving reasoning data can be effectively utilized to perform hierarchical learning with distinct emphases, as elaborated in Sec.~\ref{subsubsec:Interleaving Reasoning Generation Learning}.
Furthermore, in Sec.~\ref{subsubsec:Interleaving Reasoning Data Construction}, we introduce the data construction pipeline of \textit{\textbf{IRGL-300k}} dataset, while in Sec.~\ref{subsubsec:Inference Strategy}, we simply describe the inference strategy of IRG, such as the Classifier-Free Guidance (CFG)~\citep{ho2022classifier} condition designs.

\subsubsection{Interleaving Reasoning Generation Learning}
\label{subsubsec:Interleaving Reasoning Generation Learning}

When we set $n$ to 2, Eq.~\ref{eq2} will be reformulate as:
\begin{equation}
    \begin{split}
    & T_{in}
    \rightarrow \textcolor{red}{T_{out}^{(1)} \rightarrow I_{out}^{(1)}} \xrightarrow{\mathrm{enc}} I_{f}^{(1)}
    \rightarrow \textcolor{blue}{T_{out}^{(2)} \rightarrow I_{out}^{(2)}}.
    \end{split}
\label{eq3}
\end{equation}
It can be observed that when aiming to enhance the quality of the final image $I_{\text{out}}^{(2)}$ in IRG pipeline, we decompose the process into four progressive intermediate steps: (1) ensuring the correctness of the initial thinking process \textcolor{red}{$T_{\text{out}}^{(1)}$}, (2) improving the quality of the initial generated image \textcolor{red}{$I_{\text{out}}^{(1)}$}, (3) generating an accurate improving‑thinking step \textcolor{blue}{$T_{\text{out}}^{(2)}$} based on the first image to guide the production of a better image, and (4) synthesizing the final high‑quality image \textcolor{blue}{$I_{\text{out}}^{(2)}$} by integrating all preceding decision steps.
Evidently, as shown in Fig~\ref{fig:overview}, various decomposed learning modes can be designed to improve the model’s intermediate reasoning capacity. 
We begin with the enhancement of the initial reasoning step, \textit{i.e.}, above steps (1) and (2), which can be instantiated as the following tasks:
\begin{itemize}
    \item \textbf{Initial Thinking Understanding Learning:}
    In this task, we aim for the model to learn how to generate the correct initial thinking process given original prompt $T_{in}$ and prior image features $I_{f}^{(1)}$. 
    The design insight behind this task is as follows: when the model is provided with both a prompt and an image consistent with that prompt, we construct an auxiliary question $T_{q}$ in which the model learns, through image‑understanding supervision, how to produce a reasoning process aligned with the prompt and to recognize what kind of image such a reasoning process would yield.
    \begin{equation}
    \begin{split}
    & T_{in}+I_{f}^{(1)}+T_{q}
    \rightarrow \textcolor{red}{T_{out}^{(1)}}.
    \end{split}
    \label{eq4}
    \end{equation}
    \item \textbf{Initial Thinking Generation Learning:}
    This task directly imitates the reasoning  process for generating initial thinking given the original prompt, and can be considered a more challenging task compared to Initial Thinking Understanding Learning.
    \begin{equation}
    \begin{split}
    & T_{in}
    \rightarrow \textcolor{red}{T_{out}^{(1)}}.
    \end{split}
    \label{eq5}
    \end{equation}
    \item \textbf{Initial Full Learning:}
    In this full initial reasoning learning setting, the model learns from both text‑based reasoning sequences and high‑quality image data to enhance the quality of its initial image generation, thereby providing a stronger foundation for producing improved images in subsequent reasoning stages.
    \begin{equation}
    \begin{split}
    & T_{in}
    \rightarrow \textcolor{red}{T_{out}^{(1)} \rightarrow I_{out}^{(1)}}.
    \end{split}
    \label{eq6}
    \end{equation}
\end{itemize}
Furthermore, we also design three tasks to learning how to generate the improving thinking and improved image based on the initial reasoning step (the above steps (3) and (4)):
\begin{itemize}
    \item \textbf{Improving Thinking Understanding Learning:}
    This task is closely related to Initial Thinking Understanding Learning, but focuses on enabling the model to, given a prompt, learn to generate the improving thinking process for enhancing an initial image to an improved image.
    This is achieved by understanding the prompt $T_{in}$, comparing the differences between the features of the initial $I_{f}^{(1)}$ and improved images $I_{f}^{(2)}$ and by answering carefully designed questions $T_{q}$.
    \begin{equation}
    \begin{split}
    & T_{in}+I_{f}^{(1)}+I_{f}^{(2)}+T_{q}
    \rightarrow \textcolor{blue}{T_{out}^{(2)}}.
    \end{split}
    \label{eq7}
    \end{equation}
    \item \textbf{Improving Thinking Generation Learning:}
    This task builds on the initial reasoning stage, focusing on learning how to generate improving thinking.
    \begin{equation}
    \begin{split}
    & T_{in}+\textcolor{red}{T_{out}^{(1)}}+I_{f}^{(1)}
    \rightarrow \textcolor{blue}{T_{out}^{(2)}}.
    \end{split}
    \label{eq8}
    \end{equation}
    \item \textbf{Improving Full Learning:}
    This task represents a complete IRG process, but we constrain the model, under the condition that the initial reasoning is already completed, to learn only the improving reasoning and the high‑quality improved image components.
    As the most crucial stage of IRG, the model must learn to identify the differences in visual quality and fine‑grained fidelity between the two images, and to leverage this understanding to generate the optimal image during the improving reasoning step.
    \begin{equation}
    \begin{split}
    & T_{in}+\textcolor{red}{T_{out}^{(1)}}+I_{f}^{(1)}
    \rightarrow \textcolor{blue}{T_{out}^{(2)} \rightarrow I_{out}^{(2)}}.
    \end{split}
    \label{eq9}
    \end{equation}
\end{itemize}
The aforementioned decomposed learning modes can be clearly divided into two training objectives: (1) learning the text‑based thinking process (Initial Thinking Understanding Learning, Initial Thinking Generation Learning, Improving Thinking Understanding Learning, and Improving Thinking Generation Learning), and (2) learning the complete high‑quality image generation pipeline under the auxiliary supervision of the reasoning process (Initial Full Learning and Improving Full Learning).
This design likewise addresses the limited availability of high‑quality, full IRG thinking-image trajectories data, while learning from text‑based reasoning serves as a partial remedy to this problem.

Furthermore, we employ a two‑stage training pipeline. 
In stage 1, the model is optimized on all six tasks to generate accurate initial reasoning from a given prompt and to produce text‑based reflections derived from the initial reasoning step to enhance output quality. 
The main goal of this stage is to strengthen the text‑based reasoning capability, while incorporating full thinking–image trajectories to avoid degrading the core generative performance. 
Empirically, we find that this reasoning‑focused training converges relatively rapidly.
In the second stage, leveraging the thinking generation ability learned in Stage 1, we employ full thinking–image trajectory data (\textit{i.e.}, data in Initial Full Learning and Improving Full Learning) to efficiently optimize the entire IRG pipeline.
In this training stage involving image generation, convergence demands more iterations, since the model must spend additional training time learning the fine‑grained fidelity transformations from the initial image to the improved image.

\textbf{Discussion:}
Unfortunately, constructing complete interleaving IRG data is non‑trivial, even when considering only the two‑turn case, due to the following two challenges. 
First, obtaining final high‑quality images is inherently difficult, as the quality of existing open‑source T2I datasets remains suboptimal.
This limitation has motivated many recent works to distill images generated by powerful models such as GPT‑4o~\citep{chen2025blip3,wu2025omnigen2}. 
Second, although a subset of high‑quality data from GPT‑4o is available, IRG requires paired samples linking an initial image to its improved counterpart. 
Designing the transformation process from the initial image to the improved image is itself a challenge, which means that such pairs must be constructed from scratch and cannot be directly derived from GPT‑4o‑distilled T2I data. 
These two issues make it difficult to obtain complete IRG datasets at scale. 
To mitigate the scarcity of fully optimized training data, we design multiple intermediate training objectives. 
This is because we only learn the text-based thinking process during intermediate objective training, in order to avoid low-quality image data pollution.
We expect that, when conditions permit, access to a large quantity of complete IRG data would lead to even better performance.

\subsubsection{Interleaving Reasoning Data Construction}
\label{subsubsec:Interleaving Reasoning Data Construction}

This section introduces the IRGL-300k dataset construction pipeline for the aforementioned six decomposed learning modes.
\paragraph{Data for learning the initial reasoning step.}
For the Initial Thinking Understanding Learning task, we construct training data from open‑source T2I datasets containing prompt–image pairs. 
First, we design an initial thinking template, and then instruct a large language model (\textit{e.g.}, Qwen2.5‑VL~\citep{qwen2.5-vl}) to generate a reasoning process that is consistent with both the prompt and the corresponding image from the original T2I data.
Finally, we organize the data according to Eq.~\ref{eq4}: the prompt corresponds to $T_{\text{in}}$, the image is encoded to obtain features $I_{f}^{(1)}$, a manually designed understanding question (\textit{e.g.}, ``You have been given one prompt and one image ...'') corresponds to $T_{q}$, and the MLLM‑generated initial thinking corresponds to $\textcolor{red}{T_{\text{out}}^{(1)}}$.

For the Initial Thinking Generation Learning, the initial thinking acquisition pipeline is similar to the Initial Thinking Understanding Learning task. 
It uses the prompt and MLLM-generated initial thinking to obtain the train date (as Eq~\ref{eq5}).

For the Initial Full Learning data, to ensure learning from high‑quality image information, we input the original prompt into a high‑quality image generation model (GPT‑4o~\citep{openai2025introducing}) to produce a high‑quality image, which serves as $\textcolor{red}{I_{\text{out}}^{(1)}}$ in Eq.~\ref{eq6}. 
The initial thinking is obtained in a similar manner to that described above, by providing the prompt and the GPT‑4o‑distilled high‑quality image to an MLLM.

\paragraph{Data for learning the improving reasoning step.}
For generating data for the improving reasoning step, we encounter a key challenge: given access to a high‑quality image to serve as the improved image, we must determine the source of the initial reasoning step data.
We choose to use data generated by the base model (\textit{i.e.}, BAGEL~\citep{deng2025emerging}) conditioned on the same prompt as the initial reasoning step data. 
This design decision is motivated by two considerations: (1) it provides a simple and efficient way to obtain multi‑turn IRG data at scale, and (2) it allows us to improve the model’s performance without compromising the original capabilities of the base model.

For the Improving Thinking Understanding Learning task, we first input the prompt into the base model to generate the initial thinking and the corresponding initial image. 
We then design an instructional prompt to guide the MLLM in generating improved thinking based on the base model's generated image and the image from the T2I dataset.
The model is instructed to produce the improving thinking according to a predefined template. 
We adopt a stage‑level template~\citep{xu2024llava}, which requires the model to first generate a part of the analysis of previous generated image issues.
Then, generate the stages in the format ``Detailed Explanation of Required Improvements: ...'', ``Step‑by‑Step Modification Guidance: ...'', and ``Final Comprehensive Prompt for the Improved Image: ...''.
Finally, we organize the original prompt $T_{in}$ and initial image generated by base model $I_{f}^{(1)}$, the image in the T2I dataset $I_{f}^{(2)}$, a manually designed understanding question (\textit{e.g.}, ``You have been given one prompt and two images ...'') $T_{q}$, and the improving thinking $\textcolor{blue}{T_{out}^{(2)}}$ by Eq~\ref{eq7}.

For the Improving Thinking Generation Learning task, the data is same as Improving Thinking Understanding Learning, while it use the initial thinking generated by base model as $\textcolor{red}{T_{out}^{(1)}}$ in Eq~\ref{eq8}

In the Improving Full Learning setting, GPT‑4o is used to produce the improved image in the IRG trajectory.
Given the prompt and the initial image from the base model, GPT‑4o generates a higher‑quality, prompt‑consistent image, which we adopt as the improved image $\textcolor{blue}{I_{out}^{(2)}}$ in Eq.~\ref{eq9}.
An MLLM is then employed to produce stage‑level improving thinking detailing the transformation from the initial image to the improved image.

\subsubsection{Inference Strategy}
\label{subsubsec:Inference Strategy}

As shown in Fig~\ref{fig:overview}, the model produces a ``text–image–text–image'' trajectory. 
This poses a challenge: in conventional diffusion‑based generation models, the CFG‑conditioning design is typically straightforward, such as directly comparing the presence versus absence of the prompt. 
In contrast, for our proposed IRG under even only two-turn reasoning pipeline, particularly before generating the improved image—there are four possible condition sources to compare (\textit{i.e.}, the prompt, the initial reasoning, the initial image, and the improving reasoning). 
Therefore, a customized CFG‑conditioning strategy is required.
Based on this, we adopt the framework incorporates two complementary CFG condition schemes: (1) conditioning with versus without the image information from the initial generation, and (2) conditioning with versus without the reflection text.
In practice, we set the  the guidance scale hyper-parameters in CFG Image condition (versus without the image information) and CFG text condition (versus without the text information) to 2.0.
This strategy helps maintain high visual quality and fidelity in images produced during the improving reasoning steps, with notable benefits for generation stability.

\begin{table*}[t]
    \centering
    \scriptsize
    \caption{\textbf{Evaluation of text-to-image generation ability on GenEval benchmark.} `Gen. Only' stands for an image generation model, and `Unified' denotes a model that has both understanding and generation capabilities.
    $\dagger$ refer to the methods using MLLM rewriter.
    * means we report the reproducing results using the official Github repository and checkpoint.
    The best Overall results are \textbf{bolded}.
    }
    \resizebox{0.99\linewidth}{!}{
    \begin{tabular}{c|c|cccccc|c}
        \toprule
        \textbf{Type} & \textbf{Model}  & \textbf{Single Obj.} & \textbf{Two Obj.} & \textbf{Counting} & \textbf{Colors} & \textbf{Position} & \textbf{Color Attri.} & \textbf{Overall$\uparrow$} \\
        \midrule
        \multirow{8}{*}{\rotatebox{90}{\textit{Gen. Only}}}
        & PixArt-$\alpha$~\citep{chen2024pixart} &  0.98 & 0.50 & 0.44 & 0.80 & 0.08 & 0.07 & 0.48 \\
        & SDv$2.1$~\citep{rombach2022high} & 0.98 & 0.51 & 0.44 & 0.85 & 0.07 & 0.17 & 0.50 \\
        & DALL-E $2$~\citep{dalle2}  & 0.94 & 0.66 & 0.49 & 0.77 & 0.10 & 0.19 & 0.52 \\
        & Emu$3$-Gen ~\citep{emu3}  & 0.98 & 0.71 & 0.34 & 0.81 & 0.17 & 0.21 & 0.54 \\
        & SDXL~\citep{sdxl} &  0.98 & 0.74 & 0.39 & 0.85 & 0.15 & 0.23 & 0.55 \\
        & DALL-E $3$~\citep{dalle3} & 0.96 & 0.87 & 0.47 & 0.83 & 0.43 & 0.45 & 0.67 \\
        & SD3-Medium~\citep{SD3} & 0.99 & 0.94 & 0.72 & 0.89 & 0.33 & 0.60 & 0.74 \\
        & FLUX.1-dev$^{\dagger}$~\citep{flux} & 0.98 & 0.93 & 0.75 & 0.93 & 0.68 & 0.65 & 0.82 \\
        \midrule
        \multirow{17}{*}{\rotatebox{90}{\textit{Unified}}}
        & Chameleon~\citep{chameleon} &  - & - & - & - & - & - & 0.39 \\
        & LWM~\citep{lwm} &  0.93 & 0.41 & 0.46 & 0.79 & 0.09 & 0.15 & 0.47 \\
        & SEED-X~\citep{seed-x}  & 0.97 & 0.58 & 0.26 & 0.80 & 0.19 & 0.14 & 0.49 \\
        & TokenFlow-XL~\citep{qu2024tokenflow} &  0.95 & 0.60 & 0.41 & 0.81 & 0.16 & 0.24 & 0.55 \\
        & ILLUME~\citep{wang2024illume} &  0.99 & 0.86 & 0.45 & 0.71 & 0.39 & 0.28 & 0.61 \\
        & Janus~\citep{wu2025janus} & 0.97 & 0.68 & 0.30 & 0.84 & 0.46 & 0.42 & 0.61 \\
        & Transfusion~\citep{transfusion} & - & - & - & - & - & - & 0.63 \\
        & Emu$3$-Gen$^{\dagger}$\citep{emu3} & 0.99 & 0.81 & 0.42 & 0.80 & 0.49 & 0.45 & 0.66 \\
        & Show-o~\citep{xie2024show} &  0.98 & 0.80 & 0.66 & 0.84 & 0.31 & 0.50 & 0.68 \\
        & Janus-Pro-7B~\citep{januspro2025} &  0.99 & 0.89 & 0.59 & 0.90 & 0.79 & 0.66 & 0.80 \\
        & MetaQuery-XL$^{\dagger}$~\citep{pan2025transfer} &  -& - & - & -& -& -& 0.80 \\
        & BAGEL*~\citep{deng2025emerging} & 0.99 & 0.95  & 0.76 & 0.87 & 0.50 & 0.60 & 0.78 \\
        & Show-o2~\citep{xie2025show} &  1.00 & 0.87 & 0.58 & 0.92 & 0.52 & 0.62 & 0.76 \\
    \cmidrule{2-9}
    & BAGEL $w/$ self-CoT*~\citep{deng2025emerging} & 0.99 & 0.92  & 0.75 & 0.89 & 0.54 & 0.63 & 0.79 \\
    & \cellcolor[gray]{0.8}IRG (Ours) & \cellcolor[gray]{0.8}0.98 & \cellcolor[gray]{0.8}0.94  & \cellcolor[gray]{0.8}0.83 & \cellcolor[gray]{0.8}0.86 & \cellcolor[gray]{0.8}0.74 & \cellcolor[gray]{0.8}0.73 & \cellcolor[gray]{0.8}\textbf{0.85} \\
    \midrule
    & \color[gray]{0.4}GPT-4o~\citep{openai2025chatgpt4o} & \color[gray]{0.4}0.99 & \color[gray]{0.4}0.92 & \color[gray]{0.4}0.85 & \color[gray]{0.4}0.92 & \color[gray]{0.4}0.75 & \color[gray]{0.4}0.61 & \color[gray]{0.4}0.84 \\
    \bottomrule
    \end{tabular}
    }
    \label{tab:geneval}
\end{table*}

\begin{table*}[t]
    \centering
    \scriptsize
    \caption{\textbf{Comparison of world knowledge reasoning on WISE.} WISE examines the complex semantic understanding and world knowledge for T2I generation. `Gen. Only' stands for an image generation model, and `Unified' denotes a model that has both understanding and generation capabilities.
    * means we report the reproducing results using the official Github repository and checkpoint.
    The best results are \textbf{bolded}.
    }
    \label{tab:wisescore}
    \resizebox{0.99\linewidth}{!}{
    \begin{tabular}{c|c|cccccc|c}
    \toprule
    \textbf{Type} & \textbf{Model} & \textbf{Cultural}  & \textbf{Time}     & \textbf{Space}    & \textbf{Biology}    & \textbf{Physics} & \textbf{Chemistry} & \textbf{Overall$\uparrow$} \\
    \midrule
            \multirow{6}{*}{\rotatebox{90}{\textit{Gen. Only}}} &
 SDv1.5~\citep{rombach2022high} & 0.34 & 0.35& 0.32&0.28 &0.29 &0.21 & 0.32\\
& SDXL~\citep{sdxl} &0.43  & 0.48 &0.47  &0.44  &0.45 &0.27 & 0.43 \\
& SD3.5-large~\citep{SD3} & 0.44 &0.50 &0.58  & 0.44&0.52 &0.31 & 0.46 \\
& PixArt-Alpha~\citep{chen2024pixart} & 0.45  & 0.50& 0.48 & 0.49& 0.56 &0.34 & 0.47\\
& playground-v2.5~\citep{li2024playground} & 0.49  &0.58  & 0.55&0.43  & 0.48&0.33 & 0.49 \\
& FLUX.1-dev~\citep{flux} & 0.48  &0.58 &0.62  &0.42  &0.51 & 0.35 & 0.50 \\
    \midrule
    \multirow{10}{*}{\rotatebox{90}{\textit{Unified}}} 
& Janus~\citep{wu2025janus} &0.16 &0.26 &0.35 & 0.28 &0.30 & 0.14& 0.23\\
 &VILA-U~\citep{vila-u} & 0.26 &0.33  & 0.37 &0.35  &0.39 &0.23 & 0.31\\
& Show-o-512~\citep{xie2024show} & 0.28 &0.40  &0.48 & 0.30& 0.46 & 0.30 & 0.35\\
& Janus-Pro-7B~\citep{januspro2025} & 0.30& 0.37& 0.49 & 0.36&0.42 &0.26 & 0.35 \\
& Emu3~\citep{emu3} & 0.34 & 0.45 & 0.48 & 0.41  & 0.45 & 0.27 & 0.39 \\
& MetaQuery-XL~\citep{pan2025transfer} & 0.56& 0.55 &0.62 &  0.49 &  0.63 & 0.41 & 0.55 \\
& BAGEL~\citep{deng2025emerging} & 0.44 & 0.55 & 0.68 & 0.44 & 0.60 & 0.39 & 0.52 \\
& Show-o2*~\citep{xie2025show} & 0.64 & 0.58 & 0.61 & 0.58 & 0.63 & 0.49 & 0.61 \\
\cmidrule{2-9}
& BAGEL $w/$ self-CoT~\citep{deng2025emerging} & 0.76 & 0.69 & 0.75 & 0.65 & 0.75 & 0.58 & 0.70 \\
& \cellcolor[gray]{0.8}IRG (Ours) & \cellcolor[gray]{0.8}\textbf{0.78} & \cellcolor[gray]{0.8}\textbf{0.72} & \cellcolor[gray]{0.8}\textbf{0.76} & \cellcolor[gray]{0.8}\textbf{0.81} & \cellcolor[gray]{0.8}\textbf{0.82} & \cellcolor[gray]{0.8}\textbf{0.78} & \cellcolor[gray]{0.8}\textbf{0.77} \\
\midrule
& \color[gray]{0.4}GPT-4o~\citep{openai2025chatgpt4o} & \color[gray]{0.4}0.81 & \color[gray]{0.4}0.71 & \color[gray]{0.4}0.89 & \color[gray]{0.4}0.83 & \color[gray]{0.4}0.79 & \color[gray]{0.4}0.74 & \color[gray]{0.4}0.80 \\
        \bottomrule
    \end{tabular}
    }
\end{table*}

\begin{table}[t]
\centering
\caption{\textbf{Quantitative evaluation results of instruct-following capability on TIIF testmini (QwenVL2.5-72B as the evaluation).} 
* means we report the reproducing results using the official Github repository and checkpoint.
The best Overall and Avg. results are \textbf{bolded}.}
\renewcommand{\arraystretch}{1.7} 
\setlength{\tabcolsep}{3pt}

\centering
\resizebox{0.99\linewidth}{!}{
\begin{tabular}{c|cc|cc|cccccc|cc|cccccccccc|cc}
\toprule
\multirow{3}{*}{\textbf{Model}}
  & \multicolumn{2}{c|}{\multirow{2}{*}{\textbf{Overall}}}
  & \multicolumn{8}{c|}{\textbf{Basic Following}}
  & \multicolumn{12}{c|}{\textbf{Advanced Following}}
  & \multicolumn{2}{c}{\textbf{Designer}} \\

\cmidrule(lr){4-11} \cmidrule(lr){12-23} \cmidrule(lr){24-25}

& & &
  \multicolumn{2}{c|}{Avg}                    %
  & \multicolumn{2}{c}{Attribute}
  & \multicolumn{2}{c}{Relation}
  & \multicolumn{2}{c|}{Reasoning}
  & \multicolumn{2}{c|}{Avg}                  %
  & \multicolumn{2}{c}{\makecell{Attribute\\+Relation}}
  & \multicolumn{2}{c}{\makecell{Attribute\\+Reasoning}}
  & \multicolumn{2}{c}{\makecell{Relation\\+Reasoning}}
  & \multicolumn{2}{c}{Style}
  & \multicolumn{2}{c|}{Text}
  & \multicolumn{2}{c}{\makecell{Real\\World}} \\

& short & long &          %
  short & long &          %
  short & long &          %
  short & long &          %
  short & long &          %
  short & long &          %
  short & long &          %
  short & long &          %
  short & long &          %
  short & long &          %
  short & long &          %
  short & long            %
\\
\midrule

FLUX.1-dev~\citep{flux}  &66.24	&66.72	&74.41	&76.67	&72.50	&75.50	&78.20	&79.78	&72.52	&74.73	&60.72	&60.95	&66.76	&65.50	&61.76	&60.74	&56.60	&57.49	&63.33	&60.00	&44.49	&54.75	&74.63	&72.01 \\
FLUX.1-Pro~\citep{flux} &63.75	&63.53	&71.39	&73.57	&70.00	&68.50	&68.51	&79.97	&75.66	&72.23	&64.63	&61.42	&70.69	&72.99	&62.34	&57.27	&64.65	&57.11	&63.00	&63.00	&34.39	&36.65	&69.94	&66.78 \\
DALL-E 3~\citep{dalle3} &74.47	&72.94	&77.35	&78.40	&77.62	&75.00	&80.22	&79.67	&74.22	&80.54	&70.11	&68.45	&76.65	&75.05	&68.39	&68.07	&63.64	&59.92	&79.31	&80.00	&74.07	&75.51	&76.12	&62.69 \\
SD3.5-large~\citep{SD3} &68.69	&64.92	&73.72	&72.10	&77.50	&66.50	&74.79	&77.16	&68.87	&72.64	&65.59	&63.41	&70.85	&68.22	&65.03	&62.93	&61.03	&61.66	&56.67	&60.00	&73.30	&46.15	&70.15	&69.03 \\
PixArt-$\Sigma$~\citep{chen2024pixart} &57.46	&57.04	&67.74	&68.19	&65.50	&69.50	&74.33	&72.11	&63.40	&62.96	&56.71	&54.52	&62.47	&59.67	&57.51	&55.08	&54.84	&52.64	&76.67	&73.33	&2.71	&4.98	&63.06	&63.06 \\
Lumina-Next~\citep{zhuo2024luminanext} &46.83	&51.81	&59.62	&62.48	&49.50	&61.50	&63.30	&65.51	&66.04	&60.44	&43.72	&47.20	&47.52	&51.35	&42.65	&42.06	&44.90	&50.87	&53.33	&66.67	&2.71	&6.33	&51.49	&61.57 \\
Hunyuan-DiT~\citep{li2024hunyuandit} &49.14	&52.67	&65.39	&67.79	&59.00	&63.00	&79.89	&76.82	&57.27	&63.56	&51.61	&52.25	&62.49	&59.93	&49.14	&45.71	&49.38	&54.74	&53.33	&73.33	&0.45	&2.26	&31.34	&34.70 \\
Show-o~\citep{xie2024show} &57.34	&61.33	&69.99	&75.30	&66.50	&80.00	&76.47	&71.88	&67.00	&74.04	&58.25	&58.19	&67.21	&64.33	&54.26	&58.86	&61.38	&56.19	&46.67	&66.67	&4.98	&11.31	&71.64	&68.66 \\
LightGen~\citep{wu2025lightgen} &52.84	&46.42	&68.70	&53.99	&61.00	&52.00	&73.69	&54.52	&71.40	&50.52	&54.10	&45.76	&66.82	&48.37	&52.22	&42.93	&51.07	&50.64	&43.33	&43.33	&2.26	&10.86	&53.73	&59.70 \\
SANA 1.5~\citep{xie2025sana}   &62.57	&63.48	&73.92	&72.31	&71.50	&73.00	&82.21	&78.39	&68.04	&65.52	&60.36	&60.36	&65.65	&67.33	&56.41	&56.13	&62.20	&60.18	&66.67	&76.67	&28.51	&23.53	&61.94	&70.52 \\
Infinity~\citep{han2025infinity} &60.65	&59.66	&70.90	&71.63	&73.00	&73.00	&73.75	&74.44	&65.96	&67.44	&59.80	&57.81	&68.92	&63.78	&60.53	&56.87	&55.04	&56.81	&56.67	&73.33	&22.17	&26.70	&69.78	&61.19 \\
Janus-Pro-7B~\citep{januspro2025} &65.38	&61.10	&74.99	&73.19	&74.50	&78.00	&73.69	&70.51	&76.77	&71.04	&61.77	&56.03	&65.71	&66.48	&62.01	&55.62	&61.16	&49.34	&43.33	&70.00	&38.46	&42.08	&79.48	&73.51 \\
T2I-R1~\citep{jiang2025t2i} &67.61	&68.34	&81.14	&79.45	&80.50	&78.50	&83.09	&79.49	&79.81	&80.37	&67.38	&65.90	&69.92	&65.27	&70.10	&71.62	&68.69	&64.68	&50.00	&63.33	&32.13	&37.56	&74.25	&74.25 \\
BAGEL~\citep{deng2025emerging}* &70.97  &71.79  &78.16  &78.12  &78.00  &79.50  &80.24  &79.08  &76.25  &75.77  &68.23  &68.19  &73.37  &77.49  &64.36  &66.15  &68.92  &61.48  &80.00  &80.00  &40.72  &52.40  &76.87  &74.63  \\
MidJourney v7~\citep{midjourneyv7} &65.92	&62.43	&73.96	&74.63	&75.00	&82.00	&78.74	&78.51	&68.12	&68.55	&63.44	&62.59	&70.60	&74.03	&64.43	&59.58	&58.84	&61.34	&66.67	&33.33	&31.67	&34.39	&\textbf{79.22}	&\textbf{75.32} \\
Show-o2*~\citep{xie2025show} &62.80	&63.87	&75.30	&74.45	&73.00	&71.00	&77.22	&74.09	&75.69  &78.25	&61.38 	&66.12	&63.47  	&67.44	&62.63  &70.31	&64.15 	&60.00  &60.00	&33.33	&14.03  	&10.86	&75.00	&74.63 \\
\midrule
BAGEL $w/$ self-CoT~\citep{deng2025emerging}* &68.06 &68.78  &77.63  &79.40  &75.00  &77.00  &78.55  &82.37  &79.33  &78.81  &71.24  &68.20  &77.65  &75.37  &69.77  &65.87  &72.93  &67.91  &69.93  &63.33  &26.24  &26.70   &69.78  &71.64  \\
\cellcolor[gray]{0.8}IRG (Ours) & \cellcolor[gray]{0.8}\textbf{76.00} & \cellcolor[gray]{0.8}\textbf{73.77} & \cellcolor[gray]{0.8}\textbf{83.17} & \cellcolor[gray]{0.8}\textbf{81.28} & \cellcolor[gray]{0.8}81.00 & \cellcolor[gray]{0.8}76.00 & \cellcolor[gray]{0.8}82.96 & \cellcolor[gray]{0.8}81.86 & \cellcolor[gray]{0.8}85.54 & \cellcolor[gray]{0.8}85.98 & \cellcolor[gray]{0.8}\textbf{75.25} & \cellcolor[gray]{0.8}\textbf{74.66} & \cellcolor[gray]{0.8}75.82 & \cellcolor[gray]{0.8}77.25 & \cellcolor[gray]{0.8}78.16 & \cellcolor[gray]{0.8}77.76 & \cellcolor[gray]{0.8}73.84 & \cellcolor[gray]{0.8}72.93 & \cellcolor[gray]{0.8}90.00 & \cellcolor[gray]{0.8}70.00 & \cellcolor[gray]{0.8}43.89 & \cellcolor[gray]{0.8}47.51 & \cellcolor[gray]{0.8}72.76 & \cellcolor[gray]{0.8}74.63 \\
\midrule
\color[gray]{0.4}GPT-4o~\citep{openai2025chatgpt4o} &\color[gray]{0.4}84.19	&\color[gray]{0.4}84.61	&\color[gray]{0.4}85.30	&\color[gray]{0.4}86.55	&\color[gray]{0.4}81.00	&\color[gray]{0.4}82.12	&\color[gray]{0.4}86.16	&\color[gray]{0.4}84.12	&\color[gray]{0.4}88.74	&\color[gray]{0.4}94.50	&\color[gray]{0.4}81.24	&\color[gray]{0.4}79.75	&\color[gray]{0.4}81.95	&\color[gray]{0.4}81.55	&\color[gray]{0.4}80.03	&\color[gray]{0.4}79.85	&\color[gray]{0.4}80.88	&\color[gray]{0.4}75.68	&\color[gray]{0.4}76.67	&\color[gray]{0.4}86.67	&\color[gray]{0.4}92.76	&\color[gray]{0.4}90.05	&\color[gray]{0.4}89.55	&\color[gray]{0.4}88.06  \\
\bottomrule
\end{tabular}
}
\label{tab:tiif_qwen}
\end{table}

\begin{table}[t]
\centering
\caption{
\textbf{GenAI-Bench Evaluation Results.}
* means we report the reproducing results using the official Github repository and checkpoint.
The best results are \textbf{bolded}.
}
\label{benchmark:genai}
\resizebox{0.99\linewidth}{!}{
\begin{tabular}
{c|cccccc|cccccc|c}
\toprule
\multirow{3}{*}{\textbf{Model}} & \multicolumn{6}{c|}{\textbf{Basic Prompt}} & \multicolumn{6}{c|}{\textbf{Advanced Prompt}} & \multirow{3}{*}{\bf Overall{$\uparrow$}} \\
\cmidrule(lr){2-7} \cmidrule(lr){8-13}
 & \multirow{2}{*}{\bf Attribute{$\uparrow$}} & \multirow{2}{*}{\bf Scene{$\uparrow$}} & \multicolumn{3}{c}{\bf Relation} & \multirow{2}{*}{\bf Avg$\uparrow$} & \multirow{2}{*}{\bf Count{$\uparrow$}} & \multirow{2}{*}{\bf Differ{$\uparrow$}} & \multirow{2}{*}{\bf Compare{$\uparrow$}} & \multicolumn{2}{c}{\bf Logical} & \multirow{2}{*}{\bf Avg{$\uparrow$}}   \\
\cmidrule{4-6} \cmidrule{11-12}
&  &  & Spatial{$\uparrow$} & Action{$\uparrow$} & Part{$\uparrow$} &  & & & & Negate{$\uparrow$} & Universal{$\uparrow$} & \\
\midrule
SD v2.1 \citep{rombach2022high} & 0.80 & 0.79 & 0.76 & 0.77 & 0.80 & 0.78 & 0.68 & 0.70 & 0.68 & {0.54} & 0.64 & 0.62 & 0.70 \\
SD-XL \citep{sdxl}  & 0.84 & 0.84 & 0.82 & 0.83 & {0.89} & 0.83 & 0.71 & 0.73 & 0.69 & 0.50 & 0.66 & 0.63 & 0.73 \\
FLUX.1-dev \citep{flux}  & {0.87} & {0.88} & {0.87} & {0.85} & {0.87} & {0.87} & 0.75 & {0.78} & 0.74 & 0.45 & 0.70 & 0.64 & 0.76 \\
LWM \citep{lwm} & 0.63 & 0.62 & 0.65 & 0.63 & 0.70 & 0.63 & 0.59 & 0.58 & 0.54 & 0.49 & 0.52 & 0.53 & 0.58 \\
Show-o~\citep{xie2024show}  & 0.72 & 0.72 & 0.70 & 0.70 & 0.75 & 0.70 & 0.70  & 0.62 & 0.71 & 0.51 & 0.65 & 0.60 & 0.65 \\
VILA-U \citep{vila-u} & 0.78 & 0.78 & 0.77 & 0.78 & 0.79 & 0.76 & 0.70 & 0.71 & 0.74 & 0.53 & 0.66 & 0.64 & 0.70 \\
Liquid \citep{liquid} & -- & -- & -- & -- & -- & -- & 0.76 & 0.73 & 0.74 & 0.46 & {0.74} & 0.65 & -- \\
UniTok \citep{ma2025unitok} & -- & -- & -- & -- & -- & -- & 0.76 & 0.76 & {0.79} & 0.46 & 0.73 & 0.67 & -- \\
Mogao-7B \citep{liao2025mogao} & -- & -- & -- & -- & -- & -- & 0.77 & 0.74 & 0.77 & 0.53 & 0.71 & 0.68 & -- \\
Janus-Pro-7B~\citep{janusflow2024} & 0.85 & 0.87 & 0.85 & 0.84 & 0.85 & 0.84 & 0.73 & 0.73 & 0.71 & 0.48 & 0.65 & 0.65 & 0.75 \\
BAGEL*~\citep{deng2025emerging} & {0.89} & {0.90} & {0.89} & {0.88} & {0.89} & {0.89} & {0.77} & {0.77} & {0.79} & {0.52} & {0.71} & {0.68} & 0.79 \\
{T2I-R1}~\citep{jiang2025t2i} & {0.87} & {0.89} & {0.89} & {0.87} & {0.87} & {0.88} & {0.81} & \textbf{0.82} & {0.78} & {0.60} & {0.73} & {0.73} & 0.81 \\
{Show-o2*}~\citep{xie2025show} & {0.85} & {0.88} & {0.86} & {0.87} & {0.83} & {0.85} & {0.74} & 0.74 & {0.76} & {0.43} & {0.70} & {0.64} & 0.75 \\
\midrule
BAGEL $w/$ self-CoT*~\citep{deng2025emerging} & {0.86} & {0.88} & {0.87} & {0.87} & {0.82} & {0.86} & {0.81} & {0.78} & {0.81} & \textbf{0.66} & {0.77} & {0.75} & 0.81 \\
\cellcolor[gray]{0.8}IRG (Ours) & \cellcolor[gray]{0.8}\textbf{0.90} & \cellcolor[gray]{0.8}\textbf{0.92} & \cellcolor[gray]{0.8}\textbf{0.91} & \cellcolor[gray]{0.8}\textbf{0.90} & \cellcolor[gray]{0.8}\textbf{0.90} & \cellcolor[gray]{0.8}\textbf{0.90} & \cellcolor[gray]{0.8}\textbf{0.84} & \cellcolor[gray]{0.8}{0.78} & \cellcolor[gray]{0.8}\textbf{0.83} & \cellcolor[gray]{0.8}\textbf{0.66} & \cellcolor[gray]{0.8}\textbf{0.80} & \cellcolor[gray]{0.8}\textbf{0.77} & \cellcolor[gray]{0.8}\textbf{0.84} \\
\bottomrule
\end{tabular}
}
\end{table}

\begin{table}[t]
    \centering
    \caption{\textbf{Quantitative evaluation results on OneIG-EN.} 
    The overall score is the average of the five dimensions.
    The best results are \textbf{bolded} and the second best results are \underline{underlined}.
    }
    \resizebox{0.99\linewidth}{!}{
    \begin{tabular}{c|ccccc|c}
        \toprule
        \textbf{Model} & \textbf{Alignment}& \textbf{Text} & \textbf{Reasoning} & \textbf{Style}& \textbf{Diversity} & \textbf{Overall}$\uparrow$ \\
        \midrule
        Janus-Pro~\citep{januspro2025} & 0.553  & 0.001  &   0.139     & 0.276 & \underline{0.365} & 0.267\\
        BLIP3-o~\citep{chen2025blip3} & 0.711  & 0.013  &   0.223      & 0.361 & 0.229 & 0.307\\
        BAGEL~\citep{deng2025emerging} & 0.769  & 0.244  &   0.173    & 0.367 & 0.251& 0.361\\
        Show-o2-7B~\citep{xie2025show} & 0.817 & 0.002 & 0.226 & 0.317 & 0.177&0.308\\
        SDv1.5~\citep{rombach2022high} & 0.565 & 0.010 & 0.207 & 0.383 & \textbf{0.429} &0.319\\
        SDXL~\citep{sdxl} & 0.688 & 0.029 & 0.237 & 0.332 & 0.296 &0.316\\
        FLUX.1-dev~\citep{flux} & 0.786 & \textbf{0.523} & \textbf{0.253} & 0.368 & 0.238 & \textbf{0.434}\\
        SANA-1.5 4.8B (PAG)~\citep{xie2025sana} & 0.765 & 0.069 & 0.217 & \underline{0.401} & 0.216 &0.334\\
        Lumina-Image 2.0~\citep{qin2025lumina} & \underline{0.819} & 0.106 & \underline{0.270} & 0.354 & 0.216 & 0.353\\
        \midrule
        BAGEL $w/$ self-CoT~\citep{deng2025emerging} & 0.793 & 0.020  &   0.206    & 0.390 & 0.209 & 0.324\\
        \cellcolor[gray]{0.8}IRG (Ours) & \cellcolor[gray]{0.8}\textbf{0.839} & \cellcolor[gray]{0.8}\underline{0.377} & \cellcolor[gray]{0.8}0.239 & \cellcolor[gray]{0.8}\textbf{0.427} & \cellcolor[gray]{0.8}0.192 & \cellcolor[gray]{0.8}\underline{0.415} \\
        \midrule
        \color[gray]{0.4}GPT-4o~\citep{openai2025chatgpt4o} & \color[gray]{0.4}0.851 & \color[gray]{0.4}0.857 & \color[gray]{0.4}0.345 & \color[gray]{0.4}0.462 & \color[gray]{0.4}0.151 & \color[gray]{0.4}0.533\\
        \bottomrule
    \end{tabular}
    }
    \label{tab:oneig_en}
\end{table}

\begin{table}[t]
\centering
\caption{
\textbf{Ablation study of IRG.}
The base method is BAGEL w/ self-CoT.
``High-quality Images Training'' means that we directly use the image data of Initial Full Learning and Improving Full Learning (see Sec.~\ref{subsubsec:Interleaving Reasoning Generation Learning}) to train the base model.
``Interleaving Reasoning Generation'' indicates that using the full IRG thinking-image trajectories during training.
``Decomposed Learning Modes'' means that we train the model in decomposed learning modes with two-stage training.
}
\resizebox{0.8\linewidth}{!}{
\begin{tabular}{c|c|c|c}
\toprule
{\textbf{Model}} & {\textbf{WISE}} & {\textbf{TIIF}} & {\textbf{GenAI-Bench}} \\
\midrule
BAGEL $w/$ self-CoT~\citep{deng2025emerging} & 0.70 & 68.06/68.78 & 0.81 \\
\midrule
+ High-quality Images Training & 0.73 & 70.69/69.85 & 0.80 \\
+ Interleaving Reasoning Generation & 0.76  & 73.90/71.37 & 0.83  \\
\cellcolor[gray]{0.8}+ Decomposed Learning Modes (Ours)     & \cellcolor[gray]{0.8}0.77 & \cellcolor[gray]{0.8}76.00/73.77 & \cellcolor[gray]{0.8}0.84 \\
\bottomrule
\end{tabular}
}
\label{tab:ablation}
\end{table}

\begin{table}[t]
\centering
\caption{
\textbf{Analysis of single-turn and multi-turn IRG pipeline.}
``IRG reasoning step 1'' means that the initial image generated by IRG as the evaluation images.
}
\resizebox{0.99\linewidth}{!}{
\begin{tabular}{c|c|ccc|c|c|c}
\toprule
\multirow{3}{*}{\textbf{Model}} & \multicolumn{5}{c|}{\textbf{WISE}} & {\textbf{TIIF}} & {\textbf{GenAI-Bench}} \\
\cmidrule(lr){2-6} \cmidrule(lr){7-7} \cmidrule(lr){8-8}
& \multirow{2}{*}{\textbf{Score}} & \multicolumn{4}{c|}{\textbf{Rank score}} & \multirow{2}{*}{\textbf{Score}} & \multirow{2}{*}{\textbf{Score}} \\
\cmidrule(lr){3-6}
& & \textbf{Qwen} & \textbf{GPT-4o} & \textbf{UnifiedReward}  & \textbf{Avg.} &  & \\
\midrule
IRG reasoning step 1 & 0.79 & 29\% \textcolor{red}{$\downarrow$} & 38\% \textcolor{red}{$\downarrow$} & 43\% \textcolor{red}{$\downarrow$}  & 36.7\% \textcolor{red}{$\downarrow$} & 75.84/73.90 & 0.84 \\
\cellcolor[gray]{0.8}IRG (Ours)     & \cellcolor[gray]{0.8}0.77 & \cellcolor[gray]{0.8}71\% \textcolor{green}{$\uparrow$} & \cellcolor[gray]{0.8}62\% \textcolor{green}{$\uparrow$} & \cellcolor[gray]{0.8}57\% \textcolor{green}{$\uparrow$}  & \cellcolor[gray]{0.8}63.3\% \textcolor{green}{$\uparrow$} & \cellcolor[gray]{0.8}76.00/73.77 & \cellcolor[gray]{0.8}0.84 \\
\bottomrule
\end{tabular}
}
\label{tab:rank}
\end{table}

\section{Experiments}

\subsection{Experiment Settings}

\paragraph{IRGL-300k dataset.}
For the Initial Thinking Understanding Learning, Initial Thinking Generation Learning, Improving Thinking Understanding Learning, and Improving Thinking Generation Learning tasks, we use the open‑source T2I dataset OSP1024‑286k~\citep{lin2025uniworld}.
For each of the four tasks, we sample 50K instances from the dataset for data construction.

We employ the GPT‑4o‑distilled T2I dataset BLIP3o‑60k~\citep{chen2025blip3} in Initial Full Learning. 
For Improving Full Learning, we construct the dataset by distilling GPT‑4o with a curated set of prompts. 
Specifically, this prompt set is sourced from the training set of T2I‑compbench (following~\cite{jiang2025t2i}).
In addition, for each prompt, we use Qwen3~\citep{yang2025qwen3} to generate 1–2 complex prompt variants. 
Furthermore, following \cite{chen2025blip3}, we construct prompts from common entities and then use Qwen3 to rewrite each into 2–3 complex variants (we do not use the original entity prompts directly to obtain the GPT-4o-distilled images).
In total, this procedure produces approximately 30K samples.

We use Qwen2.5VL~\citep{qwen2.5-vl} to generate the initial thinking and improving thinking processes, while GPT-4o generates high-quality images. 

\paragraph{Training settings.}
We adopt the unified multimodal understanding and generation model BAGEL~\cite{deng2025emerging} as our base model.
In the first training stage, we train the base model for 2K steps on the six decomposed learning modes using the cross‑entropy (CE) loss and mean squared error (MSE) loss. 
We then continue training the model for 30K steps on the Initial Full Learning and Improving Full Learning tasks.

\subsection{Main Results}

To thoroughly assess the visual generation performance of our model, we conduct evaluation on a series of representative benchmarks that cover complementary aspects of controllable and knowledge-grounded generation.
Together, these benchmarks provide a comprehensive view of our model’s strengths in alignment, reasoning, stylistic control, and text rendering. 
In the following, we present detailed comparisons with SoTA baselines and highlight the improvements achieved by our proposed IRG model.

\paragraph{GenEval.}
Tab.~\ref{tab:geneval} reports the quantitative results on the GenEval~\citep{ghosh2023geneval} benchmark, which evaluates compositional T2I generation across diverse object-centric attributes such as counting, color, and spatial position.
We benchmark both generation-only models and unified understanding–generation models. 
Among generation-only approaches, FLUX.1-dev achieves the best performance with an overall score of 0.82.
Within the unified category, our proposed IRG model achieves the best overall score of 0.85, consistently outperforming all baselines across multiple sub-tasks, including challenging aspects such as counting (0.83) and position (0.74).
These results establish IRG as a new state of the art on GenEval, demonstrating strong controllability and precise compositional generation capabilities.

\paragraph{WISE.}
Tab.~\ref{tab:wisescore} reports the quantitative results on the WISE \citep{niu2025wise} benchmark, which evaluates T2I models on complex semantic understanding and world knowledge reasoning across six domains: culture, time, space, biology, physics, and chemistry. Unlike compositional tests such as GenEval, WISE focuses on knowledge-grounded generation where models must accurately reflect real-world semantics in addition to compositional control.
Among generation-only models, FLUX.1-dev achieves the best performance with an overall score of 0.50.
Within the unified category, our proposed IRG model establishes a new state of the art with an overall score of 0.77. 
It consistently outperforms prior unified models across all six domains, achieving 0.78 on cultural knowledge, 0.72 on temporal reasoning, 0.76 on spatial understanding, and above 0.80 in biology and physics. 
These results indicate that IRG not only improves general controllability but also integrates world knowledge more effectively than existing approaches, setting a new benchmark for semantic alignment in T2I generation.

\paragraph{TIIF.} 
Tab.~\ref{tab:tiif_qwen} presents the quantitative results on TIIF testmini \citep{wei2025tiifbenchdoest2imodel}, a benchmark specifically designed to evaluate T2I models' ability to interpret and accurately follow complex natural language instructions. 
The benchmark covers both basic following (attributes, relations, and reasoning) and advanced following (multi-attribute reasoning, compositional control, stylistic adherence, and textual rendering), along with a separate evaluation of designer-oriented prompts. When evaluated with QwenVL2.5-72B as the reference, our IRG model achieves the best overall performance with scores of 76.00 (short) and 73.77 (long).
Notably, IRG demonstrates consistent improvements in advanced following tasks, achieving 75.25/74.66 on average, and excelling in compositional cases such as attribute+reasoning (78.16/77.76) and relation+reasoning (73.84/72.93).

Overall, these results highlight the superior instruction-following ability of IRG, which consistently generalizes across different evaluation settings and significantly outperforms prior open-source systems, establishing it as a new state of the art in controllable T2I generation.

\paragraph{GenAI-Bench.}
Tab.~\ref{benchmark:genai} reports results on GenAI-Bench \citep{li2024genaibenchevaluatingimprovingcompositional}, which probes compositional text-to-visual generation across Basic (Attribute, Scene, and Relation: Spatial/Action/Part) and Advanced (Count, Differ, Compare, Logical: Negate/Universal) prompts. 
Our model IRG attains the best Overall score 0.84, exceeding strong baselines. 
IRG delivers the strongest overall performance on GenAI-Bench, combining robust basic grounding (attributes, scenes, relations) with improved compositional and logical generalization on advanced prompts.

\paragraph{OneIG-EN.}
Tab.~\ref{tab:oneig_en} summarizes the quantitative results on OneIG-EN \citep{chang2025oneigbenchomnidimensionalnuancedevaluation}, the English track of the OneIG-Bench benchmark that evaluates fine-grained T2I generation along five dimensions: alignment, text rendering, reasoning, style, and diversity.
The final overall score is computed as the average across these dimensions. Our IRG model establishes a new state of the art among open-source approaches with an overall score of (0.415), ranking second only to GPT-4o (0.533). 
IRG achieves the best alignment (0.839) and style (0.427) scores, and maintains balanced performance across reasoning (0.239) and text rendering (0.377).
These results indicate that IRG not only excels in faithfully aligning with user prompts but also produces aesthetically consistent outputs, demonstrating its superior general-purpose generation capability.

\subsection{Analysis of experimental results}
\label{sec:analysis}

\paragraph{Ablation study.}
Tab.~\ref{tab:ablation} isolates the contributions of data and training objectives. 
Adding high-quality image training on top of a self-CoT baseline brings moderate improvements (WISE 0.70$\rightarrow$0.73, and TIIF 68.06/68.78$\rightarrow$70.69/69.85), while when introduce the IRG pipeline (\textit{i.e.}, using the data of Initial Full Learning and Improving Full Learning) achieve the significant improvement (0.03 benchmark score improvement in both WISE and GenAI-Bench).
Furthermore, adopting IRG with six decomposed learning modes yields the largest jump (WISE 0.77, TIIF 76.00/73.77, GenAI-Bench 0.84). 
This supports our hypothesis that text-only thinking supervision (Sec.~\ref{subsubsec:Interleaving Reasoning Generation Learning}) is a data-efficient proxy for scarce full IRG thinking-image trajectories, and that mixing \emph{text-based thinking learning}, and \emph{full learning} modes provides complementary signals.

\paragraph{Analysis of single-turn and multi-turn IRG pipeline.}
As shown in Tab.~\ref{tab:rank}, two‑turn IRG and the initial generated images achieve similar benchmark scores, and on some benchmarks, initial generated image even attains slightly higher scores.
However, this does not imply that multi‑turn IRG offers no benefits. 
Our proposed IRG leverages image‑conditioned reflection to enhance visual quality and fine‑grained fidelity.

\begin{figure}[H]
    \centering
    \includegraphics[scale = 0.40]{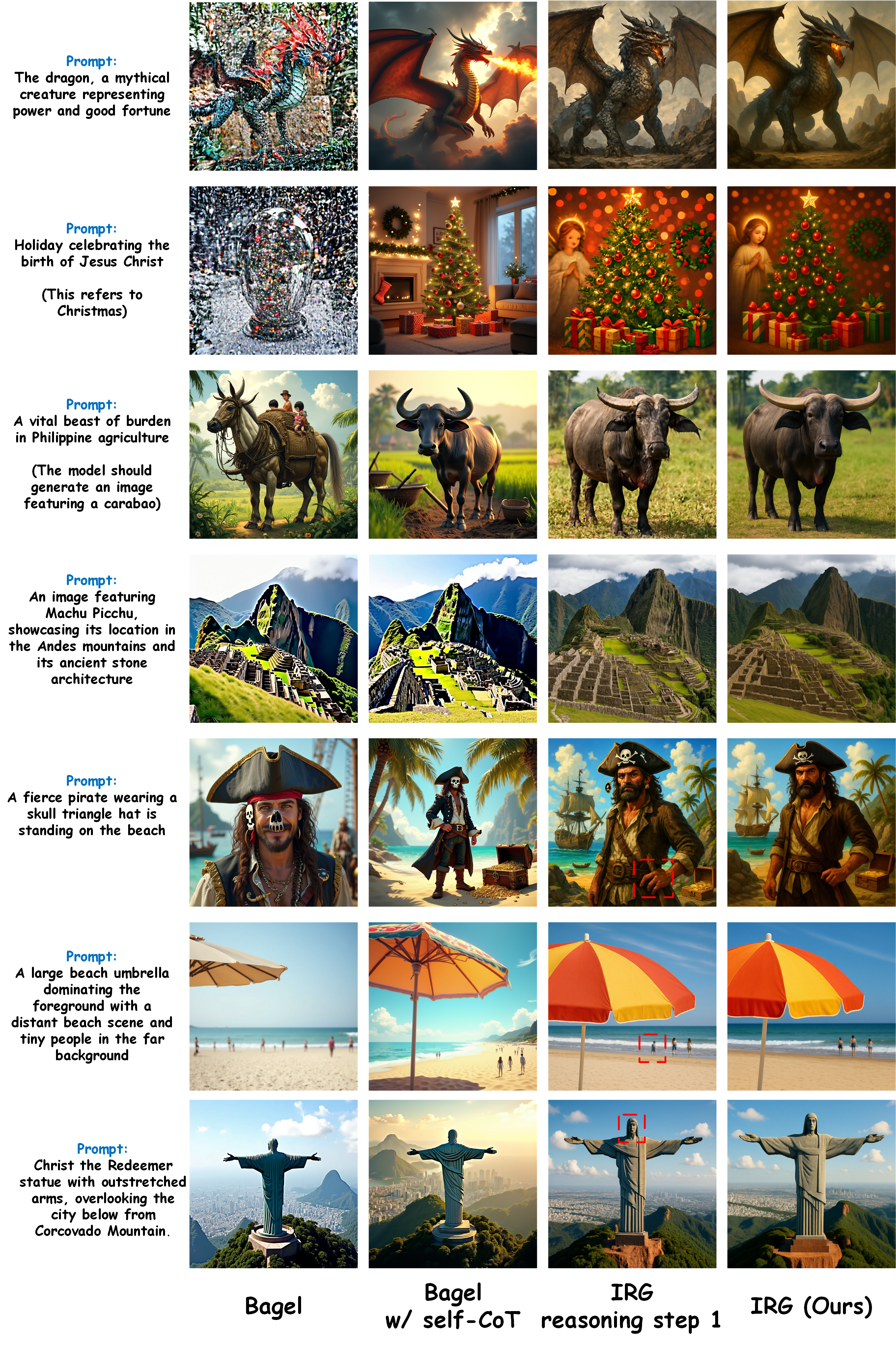}
    \caption{
        Visualization comparison results of BAGEL~\citep{deng2025emerging}, BAGEL w/ self-CoT~\citep{deng2025emerging}, IRG reasoning step 1 and our proposed IRG at 1024×1024 resolution.
        The examples are selected from WISE~\citep{niu2025wise} and GenAI-Bench~\citep{li2024genaibenchevaluatingimprovingcompositional}.
        Red boxes highlight the fine-grained details that have obvious flaws.
    }
    \label{fig:compare}
\end{figure}

We evaluate the initial generated image and improved image generated by IRG on the WISE benchmark by prompting a MLLM to directly compare the two images in terms of generation quality, fine‑grained details, aesthetics, and other visual aspects. 
To eliminate positional bias, we randomly shuffle the order of the two images presented to the MLLM and repeat the evaluation three times, reporting the averaged results. 
Multiple MLLMs are employed as evaluators, including Qwen2.5‑VL‑72B~\citep{qwen2.5-vl}, GPT‑4o~\citep{openai2025chatgpt4o}, and UnifiedReward~\citep{wang2025unified} (using its default pairwise comparison protocol, which incorporates the original task text prompt).
This multiple MLLMs as judge method mitigates evaluator-specific biases and indicates better generalization of perceived quality.

The ranking study in Tab.~\ref{tab:rank} shows that the full IRG pipeline improves agreement with multiple automatic raters compared to the first-step-only variant (average rank score: 63.3\% vs. 36.7\%), suggesting that two-turn IRG produces images whose improvements are consistently recognized by heterogeneous MLLM evaluators. 

\paragraph{Visualization comparison results.}
As illustrated in Fig.~\ref{fig:compare}, compared to BAGEL and BAGEL w/ self‑CoT, our proposed IRG achieves superior generation quality and visual fidelity. 
Moreover, relative to the first‑turn generated images, the reflection step in IRG improves the visual quality and fine‑grained fidelity of the initially generated images.
For example, enhancing suboptimal textures and refining details that were previously poorly rendered. 
This demonstrates that IRG not only produces images that are semantically correct but also places strong emphasis on fine‑detail quality in the generated content.

\paragraph{Error analysis and failure modes.}
Visualization results (Fig.~\ref{fig:case} and Fig~\ref{fig:compare}) reveal remaining failure patterns: 
(1) Micro-structure saturation on repetitive textures (\textit{e.g.}, fabrics, foliage), where the improving step occasionally over-smooths high-frequency details; 
(2) Text rendering drift under dense constraints, where the refinement trades legibility for stylistic coherence; 
(3) Global-local tension in crowded scenes, where local edits improve parts while slightly perturbing global layout. 
We find these are most pronounced when $T_{out}^{(2)}$ introduces many simultaneous edits; a conservative editing policy improves stability but may cap the attainable gains.

\section{Related Work}

\subsection{Unified Multimodal Understanding and Generation Models}

Unified Multimodal Understanding and Generation Models has attracted much attention of the research community.
The mainstream research works can be divide into three categories:
(1) Autoregressive~\citep{wu2025janus,januspro2025,lu2024unified,qu2024tokenflow,chameleon,emu3}. These methods adopt the next token prediction paradigm to generate the text and image token in one unified model.
(2) Additional Diffusion~\citep{dream-llm,wu2024next,pan2025transfer,tong2024metamorph}. They usually combine a pre-trained LLM backbone with an external diffusion module. The LLM is then used to obtain the semantic conditions that enable the diffusion module to generate an image.
And (3) Unified Integrated Transformer~\citep{deng2025emerging,liang2024mixture,janusflow2024,shi2024llamafusion,transfusion}. In this category, works typically integrate the LLM and diffusion models in one transformer model.
Our proposed IRG based on the unified integrated transformer model BAGEL~\citep{deng2025emerging} due to it was pre-trained on the large -scale interleaved text–image data.
In principle, our proposed IRG framework can be effectively applied to all of the aforementioned types of unified models, as they naturally handle both interleaved inputs and interleaved outputs.

\subsection{Reasoning Models}

Text‑based reasoning models have achieved significant progress in solving a wide range of real‑world tasks~\citep{jaech2024openai,guo2025deepseek,huang2025vision,chen2025advancing}. 
Recently, several works have begun to adopt interleaving reasoning to address more complex problems~\citep{interleaving-reasoning,openai-o3-and-o4-mini,openai-deep-research}, \textit{i.e.}, incorporating non‑text modalities into multi‑turn reasoning processes. 
In the text‑to‑image (T2I) domain, the latest studies explore whether introducing a text‑based reasoning step can enhance image generation performance~\citep{fang2025got,xiao2025mindomni,deng2025emerging,jiang2025t2i}. 
Our IRG framework is inspired by these advances and seeks to integrate interleaving reasoning into the T2I generation process.
While our approach shares conceptual similarities with reflection‑based T2I methods~\citep{zhuo2025reflection,wu2025omnigen2,chern2025thinking}, the key distinction lies in IRG’s dual focus: not only enhancing the semantic correctness of generated content but also improving visual quality, fine‑grained fidelity, and aesthetic appeal.

\section{Conclusion}

In this paper, we propose the Interleaving Reasoning Generation (IRG) framework, which generates high‑quality images through a ``text–image–text–image'' process.
Specifically, given a prompt, the model first produces a text‑based reasoning sequence and then generates an image conditioned on that reasoning. 
Next, building upon the initial image, the model reflects on how to improve its quality and produces a refined image through this reflection process.
We introduce the key designs for both the training and inference pipelines of IRG, and extensive experiments on mainstream benchmarks demonstrate significant improvements in generation performance. 
Furthermore, IRG places strong emphasis on enhancing visual quality and fine‑grained details.

\bibliography{iclr2026_conference}
\bibliographystyle{iclr2026_conference}

\end{document}